\documentclass[doublecolumn,oneside,letter]{IEEEconf}
\usepackage{amsmath}
\usepackage{dsfont}
\usepackage{amssymb}
\usepackage{graphicx}
\usepackage{subfigure}
\usepackage{verbatim}
\usepackage[thinlines,thiklines]{easybmat}
\usepackage{latexsym}
\usepackage[dvipsnames]{xcolor}
\usepackage{cite}
\usepackage{bm}
\usepackage{stmaryrd}
\usepackage{multirow}
\usepackage{textcomp}
\usepackage[ruled]{algorithm2e}
\usepackage{bookmark}
\usepackage{booktabs}

\newcommand\norm[1]{\left\lVert#1\right\rVert}

\def\s{\mathop{\rm s}\nolimits}
\def\c{\mathop{\rm c}\nolimits}

\def\diag{\mathop{\rm diag}\nolimits}
\def\rank{\mathop{\rm rank}\nolimits}
\def\spn{\mathop{\rm span}\nolimits}
\newcommand{\bs}[1]{\ensuremath{{\bm{#1}}}}

\newtheorem{lem}{Lemma}

\def\atan2{\mathrm{atan2}}

\begin{document}

\title{\Large \bf Gaussian Process-Enhanced, External and Internal Convertible (EIC) \\ Form-Based Control of Underactuated Balance Robots~\thanks{This work was partially supported by the US National Science Foundation under award CNS-1932370.}}

\author{Feng Han and Jingang Yi\thanks{F. Han and J. Yi are with the Department of Mechanical and Aerospace Engineering, Rutgers University, Piscataway, NJ 08854 USA (e-mail: {fh233@scarletmail.rutgers.edu}, {jgyi@rutgers.edu}).}}

\maketitle
\thispagestyle{empty}
\pagestyle{empty}
\begin{abstract}
External and internal convertible (EIC) form-based motion control (i.e., EIC-based control) is one of the effective approaches for underactuated balance robots. By sequentially controller design, trajectory tracking of the actuated subsystem and balance of the unactuated subsystem can be achieved simultaneously. However, with certain conditions, there exists uncontrolled robot motion under the EIC-based control. We first identify these conditions and then propose an enhanced EIC-based control with a Gaussian process data-driven robot dynamic model. Under the new enhanced EIC-based control, the stability and performance of the closed-loop system is guaranteed. We demonstrate the GP-enhanced EIC-based control experimentally using two examples of underactuated balance robots.
\end{abstract}

\section{Introduction}

An underactuated balance robot possesses fewer control inputs than the number of degrees of freedom (DOFs)~\cite{Turrisi2022RAL, KANT2020Orbital}. Control design of underactuated balance robots needs to take care of both the trajectory tracking of the actuated subsystem and balance control of the unactuated subsystem~\cite{HanTMECH2022, ChenTRO2022, HanRAL2021}. Balance of unstable coordinates of underactuated robots brings additional challenges for robot control. Many methods have been proposed to cope with the robot modeling~\cite{Turrisi2022RAL, HanRAL2021, ChenTRO2022, BECKERS2019390, ChenIROS2015}, control design and applications~\cite{GrizzleAUTO2014, Han2022Tmech}. The external and internal convertible (EIC) form-based control (i.e., EIC-based control) has been demonstrated as one of the effective approaches to achieve simultaneous trajectory tracking and balance~\cite{GetzPhD}. Other balance control algorithms include the orbital stabilization control~\cite{Maggiore2013Virtual,CANUDASDEWIT2002527, Wester2003, Grizzle2001}, and energy shaping-based control~\cite{Fantoni2000,Xin2005TAC}. One limitation of these methods is that the achieved balance-enforced trajectory is not unique~\cite{Shiriaev2005, KANT2020Orbital}.

Although the EIC-based control can achieve stability and balance~\cite{GetzPhD, HanRAL2021}, certain system conditions should be satisfied. Furthermore, an accurate robot dynamics model is required, which is also not robust under model uncertainties. Machine learning-based method provides an efficient tool in robot modeling and control. In particular, Gaussian process (GP) regression is an effective learning approach that generates analytical structure and bounded prediction errors~\cite{Lederer2023TAC, BECKERS2019390, Beckers2022TAC,pmlr2015Bach}. Development of GP-based performance-guaranteed control for underactuated balance robots has been reported~\cite{ChenTRO2022, Helwa2019RAL, Lederer2023TAC}. In~\cite{ChenTRO2022}, the control input is partitioned into two parts. A GP-based inverse dynamics controller for unactuated subsystem to achieve balance and a model predictive control (MPC) design are used to simultaneously track the given reference trajectory and obtain the balance equilibrium manifold (BEM). The GP prediction uncertainties are incorporated into the control design to enhance the control robustness. The work in~\cite{HanRAL2021} followed the cascaded control design in EIC-based framework and the controller was adaptive to the prediction uncertainties. The training data was also selected to reduce the computational complexity.

In this paper, we take advantage of the structured GP modeling in~\cite{HanRAL2021} and present a method to resolve the limitation of the original EIC-based control. We first show that under the EIC-based control, there exist uncontrolled motions that cause the entire system unstable. The uncontrolled motion is due to the fact that the EIC-based control is updated from a low- to a high-dimensional space. The conditions for the stable GP-based model learning and control are identified and presented. With the properly selected nominal model, the uncontrolled motion is eliminated with the GP-based data-driven robot dynamics. Finally, we propose a partial EIC (PEIC) form-based control by constructing a virtual inertial matrix to re-shape the dynamics coupling. The proposed GP-based control is shown to achieve guaranteed stability and performance. Experimental validation and demonstration are presented by using two examples of underactuated balance robots.

The major contributions of this work are twofold. Compared with~\cite{HanRAL2021, GetzPhD}, the uncontrolled motion of the EIC-based control are identified and illustrated. To overcome the EIC-based design limitations, the conditions for nominal GP model selection are presented. The proposed controller is new and also achieves super performance and stability. Second, unlike the work in~\cite{ChenTRO2022} with the complex MPC with high computational cost, the proposed GP models directly capture the robot dynamics and the control design preserves the EIC structure property. The demonstrated experiments are also new comparing with the previous work.


\section{EIC-Based Robot Control and Problem Statement}
\label{Sec_Model}

\subsection{Robot Dynamics and EIC-Based Control }

We consider a general underactuated balance robot with $(n+m)$ DOFs, $n, m\in \mathbb{N}$, and the generalized coordinates are denoted as $\bm q=[q_1\cdots q_{n+m}]^T$. The dynamics model can be expressed in a standard form
\begin{equation}
	\label{Eq_Physical_Model}
     \mathcal S: \bm{D}(\bm{q})\ddot{\bm{q}} + \bm{C}(\bm{q},\dot{\bm{q}})\dot{\bm{q}} + \bm{G}(\bm{q}) = \bm{B}\bm{u},
\end{equation}
where $\bm D(\bm q)$, $\bm{C}(\bm{q},\dot{\bm{q}})$ and $\bm{G}(\bm q)$ are the inertial matrix, Coriolis and gravity matrix, respectively. $\bm B$ denotes the input matrix and $\bm u \in \mathbb{R}^m$ is the control input.

The generalized coordinates are partitioned into two parts as $\bm q=[\bm q_a^T\; \bm q_u^T]^T$, with actuated and unactuated coordinates $\bm q_a \in \mathbb{R}^{n}$ and $\bm q_u \in \mathbb{R}^m$, respectively. We focus on the case $n>m$ and without loss of generality, we assume that $\bm B=[\bm I_n~\bm 0]^T$ is constant, where $\bs{I}_n$ is the identify matrix with dimension $n$. The robot dynamics~\eqref{Eq_Physical_Model} is rewritten as
\begin{subequations}
\label{Eq_Sub_Dynamics}
\begin{align}
    \mathcal S_a&: \bm D_{a a} \ddot{\bm q}_{a}+\bm D_{a u} \ddot{\bm q}_{u}+\bm H_{a}=\bm u,\label{Eq_Actuated}\\
     \mathcal S_u&:\bm D_{u a} \ddot{\bm q}_{a}+\bm D_{u u} \ddot{\bm q}_{u}+\bm H_{u}=\bm 0 \label{Eq_unactuated}
\end{align}
\end{subequations}
for actuated and unactuated subsystems, respectively. Subscripts ``$aa$ ($uu$)'' and ``$ua$ and $au$'' indicate the variables related to the actuated (unactuated) coordinates and coupling effects, respectively. For representation convenience, we introduce $\bm H=\bm{C}\dot{\bm{q}} + \bm{G}$, $\bm H_a = \bm C_{a} \dot{\bm q}+\bm G_{a}$, and $\bm H_u=\bm C_{u} \dot{\bm q}+\bm G_{u}$. The dependence of matrices $\bs{D}$, $\bm C$, and $\bm G$ on $\bm q$ and $\dot{\bm q}$ is dropped. Subsystems $\mathcal S_a$ and $\mathcal S_u$ are referred to as external and internal subsystems, respectively~\cite{GetzPhD}.

Given the desired trajectory $\bm q_a^d$ for $\mathcal S_{a}$, the control input is first designed to follow $\bm q_a^d$ by temporarily neglecting $\mathcal S_{u}$ as
\begin{equation}
\label{Eq_Actuated_Control}
\bm u^\mathrm{ext}= \bm{D}_{a a} \bm v^\mathrm{ext}+\bm{D}_{a u} \ddot{\bm q}_{u}+\bm H_{a},
\end{equation}
where error $\bm e_a=\bs{q}_a-\bm q_a^d$ and $\bm v^\mathrm{ext} \in \mathbb{R}^{n}$ is the auxiliary input such that $\bm e_a$ converges to zero. To account for the coupling relationship between $\mathcal S_{a}$ and $\mathcal S_{u}$, the unactuated coordinate $\bm q_u$ is balanced onto the BEM. The BEM is the instantaneous equilibrium in terms of $\bm q_u$ under control $\bm v^\mathrm{ext}$
\begin{equation}
\label{Eq_BEM}
\mathcal{E}=\left\{\bm q_u^{e}:  \bm \Gamma\left(\bm q_u;\bm v^\mathrm{ext}\right)=\bm 0, \dot{\bm q}_u=\ddot{\bm q}_u=\bm 0\right\},
\end{equation}
where $\bm \Gamma(\bm q_u;\bm v^\mathrm{ext})=\bm{D}_{uu} \ddot{\bm q}_u + \bm D_{ua} \bm v^\mathrm{ext}+ \bm H_{u}$. $\bm q_u^e$ is obtained by inverting $\bm \Gamma_0=\bm \Gamma(\bm q_u;\bm v^\mathrm{ext})\vert_{ \dot{\bm q}_u=\ddot{\bm q}_u=\bm 0}=\bm 0$.

To stabilize $\bm q_u$ onto $\mathcal{E}$, we update $\bm q_a$ motion to incorporate balance control as
\begin{equation}
\label{Eq_Va_Int}
\bm v^\mathrm{int}=-\bm{D}_{u a}^{+}(\bm H_{u}+\bm D_{uu} \bm v_u^\mathrm{int}),
\end{equation}
where $\bm{D}_{u a}^{+}=(\bm{D}^T_{u a}\bm{D}_{u a})^{-1}\bm{D}^T_{u a}$ denotes the generalized inverse of $\bm{D}_{u a}$. $\bm v_u^\mathrm{int}$ is the auxiliary control that drives error $\bm e_u=\bs{q}_u-\bm q_u^e$ toward zero. The final control is obtained by replacing $\bm v^\mathrm{ext}$ in~\eqref{Eq_Actuated_Control} with $\bm v^\mathrm{int}$, that is,
\begin{equation}\label{Eq_Ua_Int}
  \bm u^\mathrm{int} =  \bm{D}_{aa} \bm v^\mathrm{int}+\bm{D}_{a u} \ddot{\bm q} _{u}+\bm{H}_{a}.
\end{equation}
The above sequential EIC-based control design achieves tracking for $\mathcal S_{a}$ and balance for $\mathcal S_{u}$ simultaneously. It has been shown in~\cite{GetzPhD} that with an assumption that the robot model errors are affine with tracking errors, the control $\bm u^\mathrm{int}$ gaurantees both $\bm e_a$ and $\bm e_u$ convergence to a neighborhood of origin exponentially.

\subsection{Limitations of EIC-based Control}

In this subsection, we show the limitations of the EIC-based control design discussed in the previous section. The limitation comes from~\eqref{Eq_Va_Int} that uses a mapping from low-dimensional ($m$) to high-dimensional ($n$) space (i.e., $m<n$). For robot control~\eqref{Eq_Ua_Int}, it has been shown that there exists a finite time $T>0$ and for small number $\epsilon>0 $, $\norm{\bm q_u(t)- \bm q_u^{e}(t)}<\epsilon$ for $t>T$~\cite{GetzPhD}. Given the negligible error, we obtain $\bm D_{au}(\bm q_a, \bm q_u) \approx  \bm D_{au}(\bm q_a, \bm q_u^{e})$. We apply singular value decomposition (SVD) to $\bm D_{ua}$ and $\bm D_{ua}^+$,
\begin{align}
\label{Eq_Dua_SVD}
\bm D_{ua}=\bm U\bm \Lambda \bm V^T, \quad \bm D_{ua}^{+}=\bm V \bm \Lambda^+ \bm U^T,
\end{align}
where $\bm U \in \mathbb{R}^{m \times m}$ and $\bm V \in \mathbb{R}^{n\times n}$ are unitary orthogonal matrices. $\bm \Lambda =[\bm \Lambda_m ~\bm 0] \in\mathbb{R}^{m \times n}$ and $ \bm \Lambda^+ =[\bm \Lambda_m^{-1}~\bm 0]^T\in\mathbb{R}^{n\times m}$ and $\bm \Lambda_m=\diag(\sigma_{1},...,\sigma_{m})$ with all singular values $0 <\sigma_1\le \sigma_2 \leq \cdots \leq \sigma_{m}$.

Since $\bm V$ is a unitary orthogonal matrix, its column vectors serve as a set of complete basis in $\mathbb{R}^{n}$. Rewriting the $\bm q_a$ and $\bm v^\mathrm{ext}$ in $\spn(\bm V)$, we have transformations
\begin{equation}
\bm p_a=\bm V^T \bm q_a, \quad \bm \nu^\mathrm{ext}=\bm V^T \bm v^\mathrm{ext},
\label{transform}
\end{equation}
where $\bm \nu^\mathrm{ext}=[(\bs{\nu}_m^\mathrm{ext})^T\;(\bs{\nu}_n^\mathrm{ext})^T]^T$. Note that $[\bm p_a^T~\bm q_u^T]^T$ still serves as a complete set of generalized coordinates for $\mathcal S$. The robot dynamics $\mathcal S_u$ under control $\bm u^\mathrm{ext}$ is
\begin{equation*}
\label{Eq_S_under_uext}
  \ddot{\bm q}_u = -\bm D_{uu}^{-1}(\bm D_{ua}{\bm v}^\mathrm{ext} +\bm H_{u}).
\end{equation*}
Plugging~\eqref{Eq_Dua_SVD} and~\eqref{transform} into the above equation yields
\begin{equation}
\label{Eq_Su_in_Vu}
\ddot{\bm q}_u = -\bm D_{uu}^{-1}(\bm U\bm \Lambda \bm \nu ^\mathrm{ext}_{m} +\bm H_{u}).
\end{equation}

For $\mathcal{E}$, $\bm q_u^{e}$ is obtained by solving $\bm \Gamma_0(\bm q_u;\bm v^\mathrm{ext})=\bm 0$. With the above discussion, we substitute $\bar{\bm D}_{ua}(\bm q_u^{e})$ with $\bar{\bm D}_{ua}(\bm q_u)$ in $\bm \Gamma_0$ and therefore, using~\eqref{Eq_Dua_SVD}, $\bm \Gamma_0=\bm 0$ is rewritten into
\begin{align}\label{Eq_Gamma_BEM_Vu}
\bm \Lambda  \bm \nu_{m}^\mathrm{ext}+ \bm U^T \bm H_u^{gp}\Big\vert_{\bm q_u =\bm q_u^{e}, \dot{\bm q}_u= \ddot{\bm q}_u=\bm 0}=\bm 0,
\end{align}
which is also obtained from right-side of~\eqref{Eq_Su_in_Vu}. The BEM $\mathcal{E}$ only depends on $\bm \nu_{m}^\mathrm{ext}$, which is in the subspace $\spn\{\bm V_{1},...,\bm V_{m}\}$ of $\bs{V}$. The control effect $\bs{\nu}_n^\mathrm{ext}$ in the subspace $\ker(\bm D_{ua})$  is disposable when obtaining the BEM.

The control $\bm v^\mathrm{int}$ in~\eqref{Eq_Va_Int} is augmented by matrix $\bm D_{ua}^{+}$ using $\bm v_u^\mathrm{int}$, which is a map from a low- ($m$) to high-dimensional ($n$) space. We substitute~\eqref{Eq_Dua_SVD} into~\eqref{Eq_Va_Int} and the motion of $\mathcal S_a$ under control $\bm v_u^\mathrm{int}$ becomes
\begin{equation}
\label{Eq_Sa_Ua_Int}
\ddot{\bm q}_a=\bm v^\mathrm{int}=-\bm V \bm \Lambda^{+} \bm U^T\left(\bm H_{u}+ \bar{\bm D}_u \bm v_u^{\mathrm{int}}\right).
\end{equation}
We rewrite the above equation in the new coordinate $\bs{p}_a$ and under the EIC-based control, the closed-loop of $\mathcal S$ becomes
\begin{subequations}
\label{Eq_Su_Vu2}
\begin{align}
&\ddot p_{ai}=-\frac{\bm U_{i}^T\left(\bm H_{u}+\bm D_{uu}  \bm v_u^{\mathrm{int}}\right)}{\sigma_i},\;i=1,...,m,\label{Eq_Su_Vu2-a} \\
&\ddot p_{a(m+j)}=0, \; j=1,...,n-m,\label{Eq_Su_Vu2-b} \\
&\ddot{\bs{q}}_u = \bs{v}_u^{\mathrm{int}}.
\end{align}
\end{subequations}
Obviously, no control effect appears for coordinates in $\ker(\bm D_{ua})$ as shown by~\eqref{Eq_Su_Vu2-b} and only $m$ actuated coordinates in $\spn(\bm V)$ are under active control; see~\eqref{Eq_Su_Vu2-a}. The above results reveal the limitation of the EIC-based control design. The designed control steers only a part of the generalized coordinates and the other part is then without control.

With the above-revealed limitation of the EIC-based control and considering the data-driven model for robot dynamics, this work mainly focuses on the following problem.

\emph{Problem statement}: The goal of robot control is to design an enhanced EIC-based control to drive the actuated coordinate $\bs{q}_a$ to follow a given profile $\bs{q}_a^d$ and simultaneously the unactuated coordinate $\bs{q}_u$ to be stabilized on estimated $\mathcal{E}$ using the GP-based data-driven model.

\section{GP-Based Robot Dynamics Model}
\label{Sec_Nominal_Model}

In this section, we build a GP-based dynamics model. The enhanced EIC-based control design in the next section will be built on a selected nominal model.

\subsection{GP-Based Robot Model}

Obtaining an accurate analytical model is challenging for many robotic systems and we consider capturing the robotic system dynamics using a GP-based data-driven method. We consider a multivariate continuously smooth function $y=  f(\bm x) + w$, where $w$ is the zero-mean Gaussian noise. The Gaussian process can be viewed as a distribution over function. Denote the training data sampled from  $y= f(\bm x) + w$ is  $\mathbb D=\{\bm{X}, \bm{Y}\}=\left\{\bm  x_i, y_{i}\right\}_{i=1}^N$, where $\bm{X}=\{\bm x_i\}_{i=1}^N$, $\bm{Y}=\{y_i \} _{i=1}^N$, $\bm x_i\in \mathbb{R}^{n_x}$, and $N \in \mathbb{N}$ is the number of the data point. The GP model is trained by maximizing posterior probability $p(\bm Y; \bm X, \bm \alpha)$ over the hyperparameters $\bm \alpha$. That is, $\bm \alpha$ is obtained by solving
\begin{equation*}
  \min_{\bm \alpha} -\log(\bm Y;\bm X, \bm \alpha) =\min_{\bm \alpha} -\frac{1}{2} \bm{Y}^{T}\bm{K}^{-1} \bm{Y}  -\frac{1}{2}\log\det(\bm K),
\end{equation*}
where $\bm K=(K_{ij})$, $K_{ij}=k(\bm x_i, \bm x_j)=\sigma_{f}^{2} \exp (-\frac{1}{2}(\bm x_i-\bm x_j)^{T} \bm{W}(\bm x_i-\bm x_j))+\vartheta^2\delta_{ij}$, $\bm W = \diag\{W_1,\cdots, W_{n_x}\}>0$, $\delta_{ij}=1$ for $i=j$, and $\bm \alpha =\{\bm W, \sigma_f, \vartheta^2\}$ are hyperparameters.

Given a new $\bm  x^*$, the GP model predicts the corresponding $y$ and the joint distribution is
\begin{equation}
\begin{bmatrix} \bm Y \\  y \end{bmatrix} \sim\mathcal{N}
\left(\bm 0, \begin{bmatrix} \bm K & \bm k^T \\ \bm k & k^*   \end{bmatrix}
\right),
\end{equation}
where $\bm{k}=\bm k(\bm x^*, \bm X)$ and $k^*= k(\bm x^*, \bm x^*)$. The mean value and variance for input $\bm x^*$ are
\begin{equation}\label{Eq_GP_Pred}
    \mu_i(\bm  x^*) = \bm{k}^{T}\bm{K}^{-1}\bm Y,~ \Sigma_i(\bm x^*)=k^{*} - \bm{k}\bm{K}^{-1} \bm{k}^{T}.
\end{equation}
For a vector function, we build one GP model for each channel.

To apply the GP data-driven model for robot dynamics $\mathcal{S}$, we first build a nominal model
\begin{equation}
\label{Eq_nominal_model}
\mathcal S^n: \; \bar{\bm D}\ddot{\bm q} +\bar{\bm H}=\bm u,
\end{equation}
where $\bar{\bm D}$ and $\bar{\bm H}$ are the nominal inertia and nonlinear matrices, respectively. In general, the nominal dynamics equation does not hold for the data sampled from physical robot systems. The GP models are built to capture the difference between $\mathcal S^n$ and $\mathcal S$. The dynamics model difference is
\begin{displaymath}
\bm H^e =\bm D\ddot{\bm q}+\bm H -\bar{\bm D}\ddot{\bm q}-\bar{\bm H}=\bm u-\bar{\bm D}\ddot{\bm q}-\bar{\bm H}.
\end{displaymath}
We build the GP models to capture  $\bm H^e=[(\bm H^e_{a})^T \; (\bm H^e_{u})^T]^T$. Two GP models are built to predict $\bm H^e_a$ and $\bm H^e_u$. The training data $\mathbb D=\{\bs{X},\bs{Y}\}$ are sampled from $\mathcal S$ as $\bm{X}=\{\bm x_i\}_{i=1}^N$, $\bm{Y}=\{\bm H^e_{i} \} _{i=1}^N$, where $\bm x=\{\bm q,\,\dot {\bm q},\,\ddot {\bm q}\}$.

The GP predicted mean and variance are denoted as $(\bm \mu_i(\bm x),\bs{\Sigma}_i(\bm x))$ for $\bm H^e_i$, $i=a,u$. The GP-based robot dynamics model $\mathcal{S}^{gp}$ is then given as
\begin{subequations}
\label{Eq_gp_model}
\begin{align}
\mathcal S_a^{gp}: & \; \bar{\bm D}_{aa}\ddot{\bm q}_a +\bar{\bm D}_{au}\ddot{\bm q}_u +\bm H_a^{gp}=\bm u, \label{Eq_gp_model:a}\\
\mathcal S_u^{gp}: &\;\bar{\bm D}_{ua}\ddot{\bm q}_a+\bar{\bm D}_{uu}\ddot{\bm q}_u +\bm H_u^{gp}=\bm 0, \label{Eq_gp_model:b}
\end{align}
\end{subequations}
where $\bm H_i^{gp} =\bar{\bm H}_i+\bm \mu_i(\bm x)$, $i=a,u$. The GP-based model prediction error is
\begin{equation}
\bs{\Delta}=\begin{bmatrix} \bm \Delta_a \\ \bm \Delta_u\end{bmatrix}=\begin{bmatrix} \bm \mu_a(\bm x)-\bm H^e_a \\ \bm \mu_u(\bm x)-\bm H^e_u \end{bmatrix}.
\label{modelerror}
\end{equation}
To quantify the GP-based model prediction, we use Theorem~6 in~\cite{GPtheory} and obtain the following property for $\bm \Delta$.
\begin{lem}
\label{Lemma_GP_Errpr}
Given the training dataset $\mathbb D$, if the kernel function $k(\bm x_i, \bm x_j)$  is chosen such that $\bm H^e_a$ for $\mathcal S_a$ has a finite reproducing kernel Hilbert space norm $\norm{\bm H^e_a}_{k}<\infty$, for given $0<\eta_a<1$,
\begin{equation}\label{Eq_GP_Error}
  \Pr\left\{\norm{\bs{\Delta}_a } \leq \norm{\bm \kappa_a ^T\bm \Sigma_a ^{\frac{1}{2}}(\bm  x) }\right\} \geq \eta_a,
\end{equation}
where $\Pr\{\cdot\}$ denotes the probability of an event, $\bm \kappa_a \in \mathbb R^{n}$ and its $i$-th entry is $\kappa_{ai}=\sqrt{2\|\bs{H}^e_{a,i}\|_{k}^{2}+300 \varsigma_i \ln ^{3} \frac{N+1}{1-\eta_a^{\frac{1}{n}}}}$,  $\varsigma_i=\max_{\bm x, \bm  x^{\prime} \in \bm{X}} \frac{1}{2} \ln | 1 +\vartheta_i^{-2} k_i\left(\bm  x, \bm  x^{\prime}\right) |$. A similar conclusion holds for $\bs{\Delta}_u$ with probability $0<\eta_u<1$.
\end{lem}

\vspace{-1mm}
\subsection{Nominal Model Selection}

With the constructed GP models, the next goal is to build an enhanced EIC-based control to achieve stability and performance by eliminating the limitations that was discussed in the previous section. To achieve such a goal, we first require bounded matrices $\bar{\bs{D}}$ and $\bar{\bs{H}}$. Inverting inertial matrix $\bar{\bs{D}}$ is required for feedback linearization and thus, $\bar{\bs{D}}$ is selected invertible. Second, the uncontrolled motion exists in the kernel of matrix $\bar{\bm D}_{ua}$. If $\ker(\bar{\bm D}_{ua})$ is constant, the uncontrolled motion appears in the fixed subspace of the configuration space. Therefore, it is required that the kernel of $\bar{\bm D}_{ua}$ is non-constant. As mentioned previously, the uncontrolled motion happens due to controller updating from low- to high-dimensional spaces. If the unactuated coordinates depend on $m$ (out of $n$) control inputs, we only need to update this $m$-input set.

From the above reasoning, we obtain the following conditions for the nominal model.
\begin{itemize}
  \item $\mathcal{C}_1$: $\bar{\bm D}=\bar{\bm D}^T\succ 0$, i.e., positive definite, $\norm {\bar{\bm D}}\le d$, $\norm {\bar{\bm H}}\le h$, where constants $0< d, h < \infty $;
  \item $\mathcal{C}_2$: $\rank(\bar{\bm D}_{aa})=n$, $\rank(\bar{\bm D}_{uu})=\rank(\bar{\bm D}_{ua})=m$;
  \item $\mathcal{C}_3$: non-constant kernel of $\bar{\bm D}_{ua}$;
  \item $\mathcal{C}_4$: motion of the unactuated coordinates depend on only $m$ control inputs.
\end{itemize}
We will illustrate how to select nominal models that satisfy the above conditions in Section~\ref{Sec_Result}.

\section{GP-Enhanced EIC-Based Control}
\label{Sec_Control}

In this section, we first present the partial EIC (PEIC) control that takes advantage of the GP predictive model and explicitly eliminates the uncontrolled motion. Stability and performance analysis is then discussed.

\vspace{-1mm}
\subsection{PEIC-Based Control Design}

With GP predictive models $\mathcal S^{gp}$, we incorporate the predictive variance of $\mathcal S^{gp}_a$ into the auxiliary control $\bm v^\mathrm{ext}$ as
\begin{equation}
\hat{\bm v}^\mathrm{ext} = \ddot{\bm q}_a^d - \bm k_{p1}(\bm \Sigma_a)\bm e_a- \bm k_{d1}(\bm \Sigma_a)\dot{\bm e}_a
\label{control1}
\end{equation}
where $\bm k_{p1}(\bm \Sigma_a), \bm k_{d1}(\bm \Sigma_a)\succ0$ are control gains that depend on variance $\bm \Sigma_a$. Given the GP-based dynamics, the BEM is estimated by solving the optimization problem
\begin{equation}
\label{Eq_BEM_opt}
\hat{\bm q}_u^e=\arg \min_{\bm q_u} \| \bs{\Gamma}_0 (\bm q_u;\hat{\bm v}^\mathrm{ext})\|.
\end{equation}
The solution is denoted as $\hat{\bm q}_u^e$. The updated control design is
\begin{equation}
    \hat{\bm v}_u^\mathrm{int} = \ddot {\hat{\bm q}}_u^{e} -\bm k_{p2}(\bm \Sigma_u)\hat{\bm e}_u - \bm k_{d2}(\bm \Sigma_u)\dot{\hat{\bm e}}_u,
\label{control2}
\end{equation}
where $\hat{\bm e}_u =\bm q_u- \hat{\bm q}_u^{e}$ is the internal system tracking error relative to the estimated BEM. $\bm k_{p2}(\bm \Sigma_u), \bm k_{d2}(\bm \Sigma_u)\succ0$ are also designed and tuned by the estimated GP variance $\bm \Sigma_u$.

Let $\Delta \bm q_u^e=\bm q_u^e-\hat{\bm q}_u^e$ denote the BEM estimation error and the actual BEM is $\bm q_u^e =\hat{\bm q}_u^ {e}+\Delta \bm q_u^e$. The control design based on actual BEM is $\bm v_u^\mathrm{int} = \ddot {\bm q}_u^e -\bm k_{p2}(\bm \Sigma_u)\bm e_u - \bm k_{d2}(\bm \Sigma_u)\bm e_u$ and therefore, we have
\begin{equation*}
\bm v_u^\mathrm{int}=\hat{\bm v}_u^{\mathrm{int}}-\Delta \bm v_u^\mathrm{int},
\end{equation*}
where $\Delta \bm v_u^\mathrm{int} = \Delta \ddot{\bm q}^e_u+\bm k_{p2}\Delta \bm q^e_u+ \bm k_{d2}\Delta \dot{\bm q}^e_u$. Compared to~\eqref{Eq_BEM}, the BEM estimation error comes from GP modeling error and optimization accuracy. It is reasonable to assume that $\Delta \bm q_u^e$ is bounded. Because of bounded Gaussian kernel function, the GP prediction variances are also bounded, i.e.,
\begin{equation}
\norm{\bm{\Sigma}_a(\bm{x})} \leq (\sigma^{\max}_{a})^2,
\norm{\bm{\Sigma}_u(\bm{x})} \leq (\sigma^{\max}_{u})^2,
\label{bound}
\end{equation}
where $\sigma^{\max}_{a}=\max _{i}(\sigma_{f_{ai}}^2+\vartheta^2_{ai})^{1/2}$, $\sigma^{\max}_{u}=\max _{i}(\sigma_{f_{ui}}^2+\vartheta^2_{ui})^{1/2}$, $\sigma_{f}$ and $\vartheta$ are the hyperparameters in each channel. Furthermore, we require the control gains to satisfy the following bounds
\begin{align*}
&k_{i1}\le \lambda(\bm k_{i1})\le k_{i3}, \quad k_{i2} \le  \lambda(\bm k_{i2}) \le k_{i4}, \; i=p,d
\end{align*}
for constants $k_{pj},k_{dj}>0$, $j=1,\cdots,4$, where $\lambda(\cdot)$ denotes the eigenvalue operator.

The control design $\bm v^\mathrm{int}$ in~\eqref{Eq_Va_Int} revises the preliminary control $\bm v^\mathrm{ext}$. Under the updated control, $\bm q_a$ serves as a control input to drive $\bm q_u$ to $\bm q_u^e$. For PEIC-based control, we instead consider a partial coupling constraint between $\bm q_a$ and $\bm q_u$ and assign $m$ control inputs (equivalently the actuated coordinates) for unactuated subsystem control. To achieve such a goal, we partition the actuated coordinates as $\bm q_a=[\bm q_{aa}^T~\bm q_{au}^T]^T$, $\bm q_{au} \in \mathbb{R}^m$, $\bm q_{aa} \in \mathbb{R}^{n-m}$, and $\bs{u}=[\bs{u}_a^T \; \bs{u}_u^T]^T$. The $\mathcal S^{gp}$ dynamics in~\eqref{Eq_gp_model} is rewritten as
\begin{equation}\label{Eq_dyna1}
\begin{bmatrix}
 \bar{\bm D}_{a a}^a & \bar{\bm D}_{a a}^{au} & \bar{\bm D}_{a u}^a \\
 \bar{\bm D}_{a a}^{ua} &\bar{\bm D}_{a a}^{u} & \bar{\bm D}_{a u}^u\\
 \bar{\bm D}_{u a}^a & \bar{\bm D}_{u a}^u & \bar{\bm D}_{u u}
\end{bmatrix}
\begin{bmatrix}
  \ddot{\bm q}_{aa} \\
  \ddot{\bm q}_{au}\\
  \ddot{\bm q}_u
\end{bmatrix}
+\begin{bmatrix}
  \bm{H}_{aa}^{gp} \\
  \bm{H}_{au}^{gp} \\
  \bm H_{u}^{gp}
\end{bmatrix}
=\begin{bmatrix}
  \bm u_{a} \\
  \bm u_{u} \\
  \bm 0
\end{bmatrix},
\end{equation}
where all block matrices are in proper dimension. We rewrite~\eqref{Eq_dyna1} into three groups as
\begin{subequations}\label{Eq_S1}
\begin{align}
    \mathcal S^{gp}_{aa}&: \bar{\bm D}_{a a}^a \ddot{\bm q}_{aa}+\bm H_{an}^a=\bm u_{a},\label{Eq_Saa}\\
    \mathcal S^{gp}_{au}&:\bar{\bm D}_{a a}^{u} \ddot{\bm q}_{au}+\bar{\bm D}_{au}^u \ddot{\bm q}_{u}+\bm H_{an}^u=\bm u_{u},\label{Eq_Sau}\\
    \mathcal S^{gp}_{u}&:\bar{\bm D}_{u a}^u \ddot{\bm q}_{au}+\bar{\bm D}_{u u} \ddot{\bm q}_{u}+\bm H_{un}=\bm 0,\label{Eq_Su}
\end{align}
\end{subequations}
where $\bm H_{an}^{aa}= \bar{\bm D}_{a a}^{au} \ddot{\bm q}_{au}+ \bar{\bm D}_{a u}^{a} \ddot{\bm q}_u +\bm{H}_{aa}^{gp}$, $\bm H_{an}^u=\bar{\bm D}_{a a}^{ua} \ddot{\bm q}_{aa}+ \bar{\bm D}_{a u}^{u} \ddot{\bm q}_u +\bm{H}_{au}^{gp}$, and $\bm H_{un}=\bar{\bm D}_{u a}^a \ddot{\bm q}_{aa}+ \bm H_u^{gp}$. Apparently, $\mathcal S^{gp}_{u}$ is virtually independent of $\mathcal S^{gp}_{aa}$, since there is "no dynamics coupling". The dynamics coupling virtually exists only between $\mathcal S^{gp}_{u}$ and $\mathcal S^{gp}_{au}$.

\begin{figure}[ht!]
	\centering
	\includegraphics[height=5.8cm]{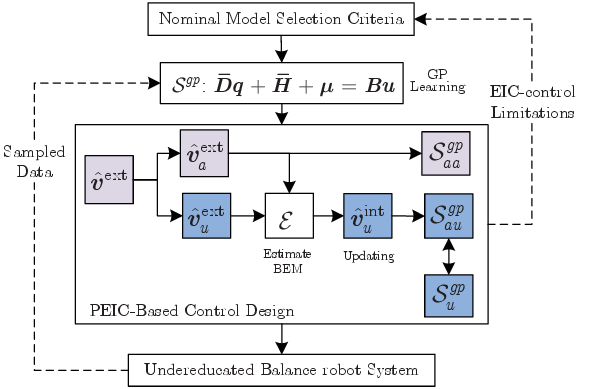}
	\caption{An overall flowchart of the PEIC-based control design.}
    \label{Fig_PEIC_Schematics}
    \vspace{-8mm}
\end{figure}

Let control $\hat{\bm v}^\mathrm{ext}$ in~\eqref{control1} be partitioned into $\hat{\bm v}_{a}^\mathrm{ext}, \hat{\bm v}_{u}^\mathrm{ext}$ corresponding to $\bm q_{aa}$ and $\bm q_{au}$, respectively. $\hat{\bm v}_{a}^\mathrm{ext}$ is directly applied to $\mathcal S^{gp}$ and $\hat{\bm v}_{u}^\mathrm{ext}$ is updated for balance control purpose. As aforementioned, the necessary conditions to eliminate the uncontrolled motion in $\mathcal S_a$ is that $\bm q_u$ only depends on $m$ inputs. The task of driving $\bm q_u$ to $\bm q_u^e$ is assigned to $\bm q_{au}$ coordinates only. With this observation, the PEIC-based control is given as $\hat{\bm u}^\mathrm{int} =[\hat{\bm u}_{a}^T~\hat{\bm u}_{u}^T]^T$ with
\begin{equation}
\label{Eq_U_PEIC}
\hat{\bm u}_{a}=\bar{\bm D}_{a a}^a \hat{\bm v}_{a}^{\mathrm{ext}}+\bm H_{a n}^a, \hat{\bm u}_{u}=\bar{\bm D}_{a a}^u \hat{\bm v}^{\mathrm{int}}+\bar{\bm D}_{a u}^u \ddot{\bm q}_u+\bm H_{a n}^u,
\end{equation}
where $\hat{\bm v}^\mathrm{int} = -\left(\bar{\bm{D}}_{u a}^u\right)^{-1}\left(\bm H_{un}+\bar{\bm D}_{uu} \hat{\bm v}_u^\mathrm{int}\right)$. The auxiliary controls are $\hat{\bm v}_a^\mathrm{ext}$ and $\hat{\bm v}_u^\mathrm{int}$. Clearly, the unactuated subsystem only depends on $\bm u_{u}$ under the PEIC design. Fig.~\ref{Fig_PEIC_Schematics} illustrates the overall flowchart of the PEIC-based control design for underactuated balance robots.

\subsection{Stability and Performance Analysis}
\label{Sec_Stability}

To investigate the closed-loop dynamics, we take the GP prediction error and BEM estimation error into consideration. The GP prediction error in~\eqref{modelerror} is extended to $\bs{\Delta}_{aa}$, $\bs{\Delta}_{au}$ and $\bs{\Delta}_u$ for $\bm q_{aa}, \bm q_{au}, \bm q_u$ dynamics, respectively. Under the PEIC-based control, the dynamics of $\mathcal S$ becomes
\begin{align*}
\ddot{\bm q}_{aa}&=\hat{\bm v}_{a}^\mathrm{ext}-(\bar{\bm D}_{aa}^a)^{-1}\bs{\Delta}_{aa},\\
\ddot{\bm q}_{au}&= -(\bar{\bm D}_{u a}^u)^{-1}(\bm H_{u n}+\bar{\bm D}_{uu} \hat{\bm v}_u^\mathrm{int})-(\bar{\bm D}_{a a}^u)^{-1} \bs{\Delta}_{au},\\
\ddot{\bm q}_{u} &=\hat{\bm v}_u^\mathrm{int}-\bar{\bm D}_{uu}^{-1}[\bs{\Delta}_u-\bar{\bm D}_{ua}^u(\bar{\bm D}_{aa}^u)^{-1} \bs{\Delta}_{au}].
\end{align*}
The BEM obtained by~\eqref{Eq_BEM_opt} under input $[\ddot{\bm q}_{aa}~\hat{\bm v}_{u}^\mathrm{ext}]$ is equivalent to inverting~\eqref{Eq_Su}. Therefore,
\begin{equation*}
\hat{\bm v}_{u}^{\mathrm{ext}}=-\left(\bar{\bm D}_{u a}^u\right)^{-1} \bm{H}_{un}\big\vert_{\substack{\bm q_u=\hat{\bm q}_u^{e}, \dot{\bm q}_u=\ddot{\bm q}_u=\bm 0}}.
\end{equation*}
Substituting the above equation into the $\bm q_{au}$ dynamics yields $\ddot{\bm q}_{au}= \hat{\bm v}_{u}^{\mathrm{ext}}+ \bm O_{au}$, where $\bm O_{au} =-(\bar{\bm D}_{u a}^u)^{-1}\bar{\bm D}_{uu} \hat{\bm v}_u^\mathrm{int}-(\bar{\bm D}_{a a}^u)^{-1} \bs{\Delta}_{au}+\bs{O}_{\text{hot}}$ and $\bs{O}_{\text{hot}}$ denotes the higher order terms.

Defining the total error $\bs{e}_q=[\bs{e}_a^T \; \bs{e}_u^T]^T$ and $\bs{e}=[\bs{e}_q^T \; \dot{\bs{e}}_q^T]^T$, the closed-loop error dynamics becomes
\begin{equation}
\label{Eq_error_dyna}
\hspace{-2mm} \dot{\bm e}=\begin{bmatrix} \dot{\bm e}_q \\ \ddot{\bm e}_q \end{bmatrix}=\underbrace{\begin{bmatrix} \bm 0 & \bs{I}_{n+m} \\ -\bs{k}_{p} & -\bs{k}_d \end{bmatrix}}_{\bm A} \begin{bmatrix} \bm e_q \\ \dot{\bm e}_q \end{bmatrix}+ \underbrace{\begin{bmatrix} \bm 0 \\ \bs{O}_\mathrm{tot} \end{bmatrix}}_{\bm O}=\bm A\bm e + \bm O
\end{equation}
with $\bm O_\mathrm{tot} = [\bm O_{a}^T~\bm O_u^T]^T$, $\bm O_a=[\bm O_{aa}^T~\bm O_{au}^T]^T$, $\bm O_{aa} =-(\bar{\bm{D}}_{a a}^a)^{-1}\bs{\Delta}_{aa}$, $ \bm O_u=-\bar{\bm{D}}_{uu}^{-1}(\bs{\Delta}_u-\bar{\bm{D}}_{u a}^u(\bar{\bm{D}}_{a a}^u)^{-1} \bs{\Delta}_{au})-\bs{\Delta} \bm v_u^\mathrm{int}$, $\bs{k}_{p}=\diag(\bs{k}_{p1},\bs{k}_{p2})$, and $\bs{k}_{d}=\diag(\bs{k}_{d1},\bs{k}_{d2})$.

Because of bounded $\bar{\bm D}$, there exists constants $0<d_{a1}, d_{a2}, d_{u1}, d_{u2}<\infty$ such that $d_{a1} \le \norm{\bar{\bm D}_{aa}}\le d_{a2}$ and $d_{u1}\le \norm{\bar{\bm D}_{uu}}\le d_{u2}$. The perturbation terms are further expressed and bounded as
\begin{align*}
\norm{\bm{\bm O}_a}& =\norm{-\begin{bmatrix}
\bm 0 \\
(\bar{\bm D}_{u a}^u)^{-1}\bar{\bm D}_{uu} \hat{\bm{v}}_u^{\mathrm{int}}
\end{bmatrix}-
(\bar{\bm D}_{a a}^a)^{-1}\bs{\Delta}_a +\begin{bmatrix} \bm 0 \\ \bm{O}_{\text{hot}}
\end{bmatrix}}\nonumber \\
& \leq \tfrac{d_{u 2}}{\sigma_1}\norm{\hat{\bm{v}}_u^{\mathrm{int}}}+\tfrac{1}{d_{a 1}}\norm{\bs{\Delta}_a}+\norm{\bm{O}_{\text{hot}}}
\end{align*}
and
\begin{align*}
\norm{\bm{O}_u}=&\norm{ -\bar{\bm D}_{u u}^{-1}(\bs{\Delta}_u-\bar{\bm D}_{u a}^u(\bar{\bm D}_{a a}^u)^{-1} \bs{\Delta}_{au})-\Delta \bm v_u^{\mathrm{int}}}\nonumber\\
& \leq \tfrac{1}{d_{u 1}}\norm{\bs{\Delta}_u}+\tfrac{\sigma_m}{d_{u 1} d_{a 1}}\norm{\bs{\Delta}_a}+\norm{\Delta \bm v_u^\mathrm{int}}.
\end{align*}
The perturbation $\bm{O}_{\text{hot}}$ is due to approximation and $\Delta \bs{v}_u^{\mathrm{int}}$ is the control difference due to the BEM calculation by the GP prediction, and we assume they are affine functions with total error $\bs{e}$, that is,
\begin{align*}
\norm{\bm{O}_{\text{hot}}}  \le  {c_1}\norm{\bm e} + c_2,\quad
\norm{\Delta\bm v_u^{\mathrm {int}}}  \leq c_3\norm{\bm e}+c_4
\end{align*}
with $0 < c_i < \infty$, $i=1,\cdots,4$. From~\eqref{bound}, we have $\|\bm \kappa_a^T \bm \Sigma_a^{\frac{1}{2}} \|\leq \sigma^{\max}_{a}\norm{\bm \kappa_a}$ and $\|\bm \kappa_u^T \bm \Sigma_u^{\frac{1}{2}}\|\leq \sigma^{\max}_{u}\norm{\bm \kappa_u}$. Thus, for $0<\eta <1$, we can show that
\begin{align*}
\Pr&\left\{\norm{\bm O}\le d_1 +d_2\norm{\bm e}+l_{u}\norm{\bm \kappa_u}+ l_{a}\norm{\bm \kappa_a }\right\}\ge \eta,
\end{align*}
with $\eta=\eta_a \eta_u$, $d_1=c_2+\left(1+\frac{d_{u 2}}{\sigma_1}\right) c_4$, $d_2=c_1+\frac{d_{u 2}}{\sigma_1} c_3$, $l_{a}=\frac{\sigma^{\max}_{a}(d_{u 1}+\sigma_m)}{d_{u1} d_{a 1}}$, $l_{u}=\frac{\sigma^{\max}_{u}}{d_{u1}}$.

With the above results, we have the following results about the stability and performance of the PEIC-based control and the proof is neglected due to page limit.
\begin{lem}
\label{lemma2}
For robot dynamics~\eqref{Eq_Sub_Dynamics}, using the GP-based model~\eqref{Eq_gp_model} and under the PEIC-based control design~\eqref{control1},~\eqref{control2} and~\eqref{Eq_U_PEIC}, the system error $\bs{e}$ exponentially converges to a small ball near the origin.
\end{lem}

\section{Experimental Results}
\label{Sec_Result}

\begin{figure}[ht!]
	\centering
	\subfigure[]{
	\label{Fig_pen_photo}
	\includegraphics[height=4.2cm]{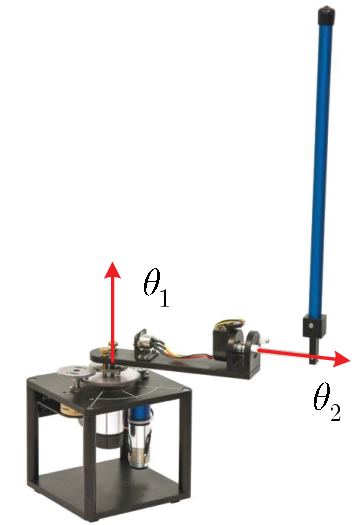}}
	\hspace{5mm}
	\subfigure[]{
	\label{Fig_Leg_photo}
	\includegraphics[height=4.2cm]{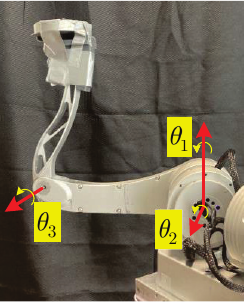}}
	\caption{(a) A Furuta pendulum testbed. The base link joint $\theta_1$ is actuated and the pendulum link joint  $\theta_2$ is unactuated. (b) A three-link robotic leg with two base links $\theta_1$ and $\theta_2$ are actuated and the top link $\theta_3$ is unactuated.}
\vspace{-3mm}
\end{figure}

We used two inverted pendulum platforms to conduct experiments to validate and demonstrate the robot control design. Fig.~\ref{Fig_pen_photo} shows a 2-DOF rotary inverted pendulum and Fig.~\ref{Fig_Leg_photo} for a 3-DOF robotic leg that has an inverted link as the controlled balance task.

\begin{figure*}[h!]
	\centering
	\subfigure[]{
		\label{Fig_pen_arm}
		\includegraphics[height=4.0cm]{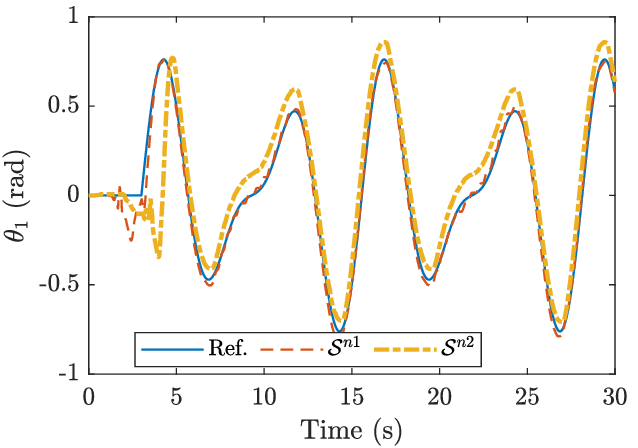}}
	\hspace{-3.0mm}
	\subfigure[]{
		\label{Fig_pen_pen}
		\includegraphics[height=4.0cm]{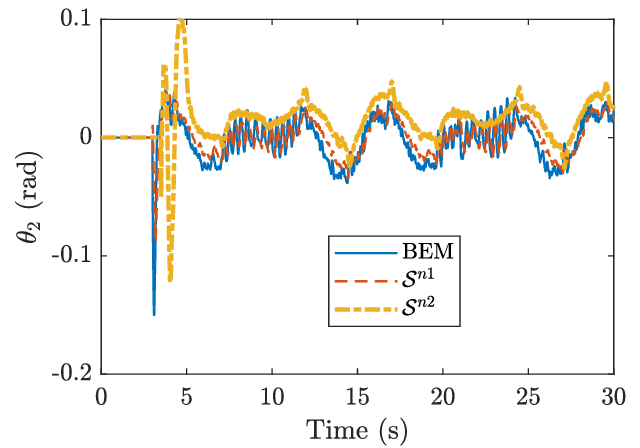}}
	\hspace{-3mm}
	\subfigure[]{
		\label{Fig_pen_error}
		\includegraphics[height=4.0cm]{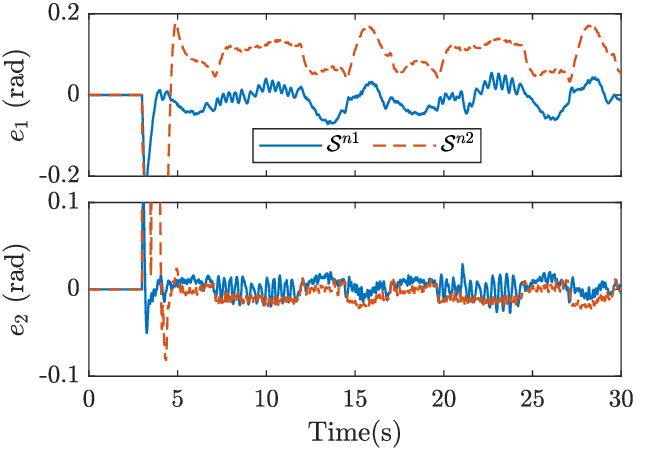}}
    \vspace{-2mm}
	\caption{Experiment results with rotary inverted pendulum (a) Arm rotation angles. (b) Pendulum rotation angles. (c) Tracking control errors.}
\vspace{-5mm}
	\label{Fig_pendulum}
\end{figure*}

The rotary inverted pendulum (2 DOFs, $n=m=1$) was made by Quanser Inc. and we used this example to illustrate the EIC-based control. The base joint ($\theta_1$) is actuated by a DC motor and the inverted pendulum joint ($\theta_2$) is unactuated. The physical model in~\eqref{Eq_Sub_Dynamics} is given in~\cite{Apk2011}. The control input is motor voltage. Since the condition $\mathcal C_4$ is satisfied automatically, there is no uncontrolled motion if the EIC-based control is applied. Either a constant nominal model or a time-varying nominal model should work. We take the nominal model
\begin{align*}
\mathcal S^{n1}:& \;\; \bar{\bm D}_1=\frac{1}{100}\begin{bmatrix}
		5 & -2\c_{2}\\
		-2\c_{2} & 2
	\end{bmatrix},~\bar{\bm H}_1=\begin{bmatrix}
		0 \\
		-\s_2
	\end{bmatrix}, \\
\mathcal S^{n2}: & \;\; \bar{\bm D}_2=\frac{1}{100}\begin{bmatrix}
		2 & 1\\
		1 & 2
	\end{bmatrix},~\bar{\bm H}_2=\bm 0,
\end{align*}
where $\c_i=\cos\theta_i$, $\s_i=\sin\theta_i$ for angle $\theta_i$, $i=1,2$. The control gains $k_{p1}=10+50\Sigma_a$, $k_{d1}=3+10\Sigma_a$, $k_{p2}=1000+500\Sigma_u$, and $k_{d2}=100+200\Sigma_u$ were chosen. The reference trajectory was $\theta_1=0.5\sin t+0.3\sin 1.5t $~rad. The control was implemented at $400$~Hz in Matlab/Simulink with Quanser's hardware-in-the-loop real-time system. For comparison purpose, we also implemented a physical model-based EIC controller in experiments.

Fig.~\ref{Fig_pendulum} shows the experimental results. With either $\mathcal S^{n1}$ or $\mathcal S^{n2}$, the base link closely follows the reference trajectory and a similar trend is found for the pendulum motion (see Fig.~\ref{Fig_pen_pen}). However, the tracking error was reduced and the pendulum closely followed the small vibrations for the case with $S^{n1}$. With $\mathcal S^{n2}$, the tracking errors became large when the base link changed rotation direction; see Fig.~\ref{Fig_pen_error} at $t=10, 17, 22$~s. Since the condition $\mathcal C_4$ is automatically satisfied, both the time-varying nominal model and constant nominal model worked for modeling learning and EIC-based control design. Table~\ref{Table_pen_eror} lists the statistics of the tracking errors (mean and one standard deviation), including the learning-based control and physical model-based control. For both subsystems, the errors with the learning-based approach are smaller. In particular, with a time-varying nominal model, the tracking error (mean value) for ${e}_1$ and $e_2$ reduced $75\%$ and $65\%$ respectively in comparison with the physical model-based one.

\renewcommand{\arraystretch}{1.3}
\setlength{\tabcolsep}{0.1in}
\begin{table}[ht!]
  \centering
    \caption{Tracking Errors of Rotatory Inverted Pendulum}
    \vspace{-2mm}
    \label{Table_pen_eror}
  \begin{tabular}{|c|c|c|c|}
    \hline\hline
       & $\mathcal S^{n1}$ & $\mathcal S^{n2}$ & Model-Based\\ \hline
    $|e_1|$ (rad) & $0.024\pm 0.017$ & $0.081\pm 0.105$ & $0.109\pm 0.040$ \\ \hline
    $|e_2|$ (rad) & $0.009\pm0.005$ & $0.009\pm 0.008$ & $0.026\pm 0.015$\\
    \hline\hline
  \end{tabular}
\end{table}

\begin{figure*}[h!]
	\centering
	\subfigure[]{
		\label{Fig_Leg_Traj}
		\includegraphics[height=4.5cm]{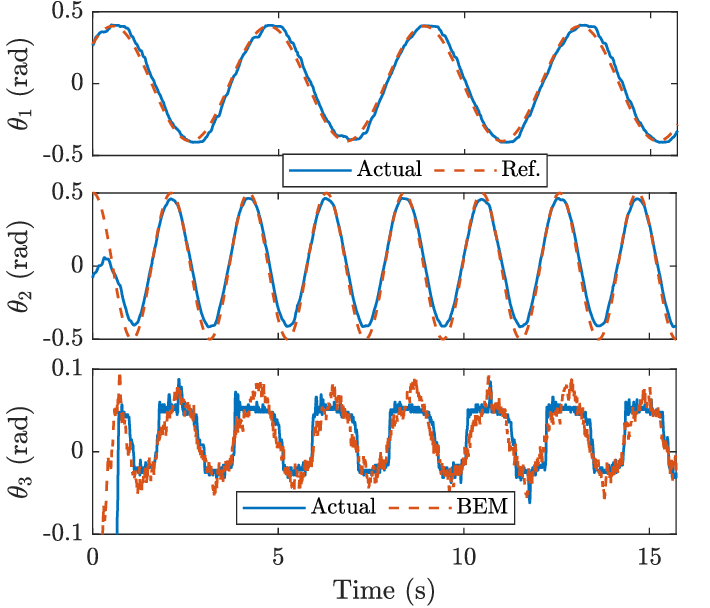}}
	\hspace{-1mm}
	\subfigure[]{
		\label{Fig_Leg_Error}
		\includegraphics[height=4.5cm]{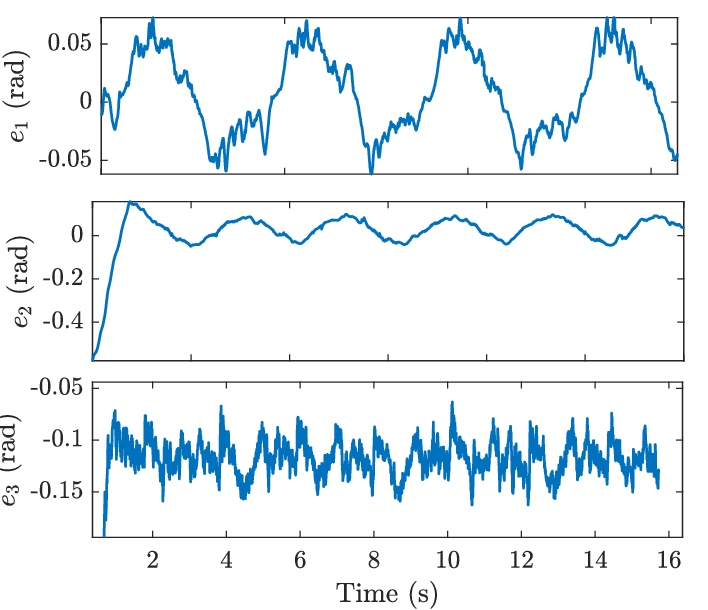}}
	\hspace{-1mm}
	\subfigure[]{
		\label{Fig_Leg_Traj_EIC}
		\includegraphics[height=4.5cm]{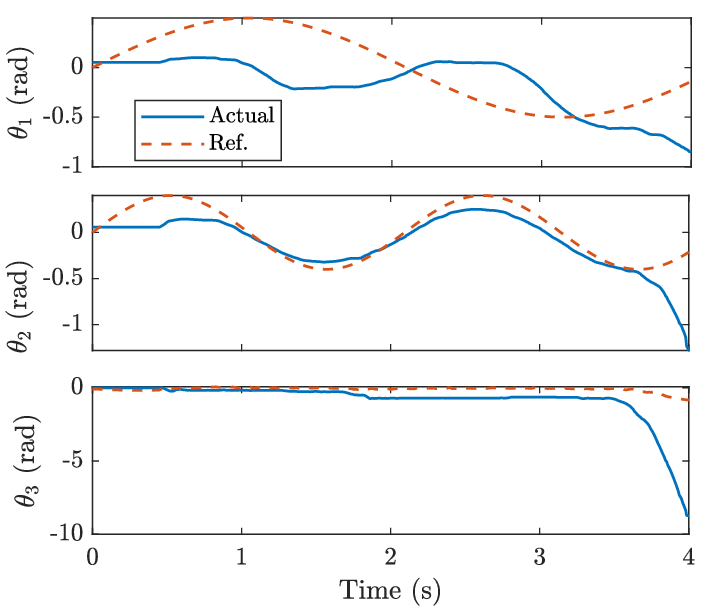}}
	\vspace{-1mm}
	\subfigure[]{
		\label{Fig_Leg_Traj_P}
		\includegraphics[width=5.35cm]{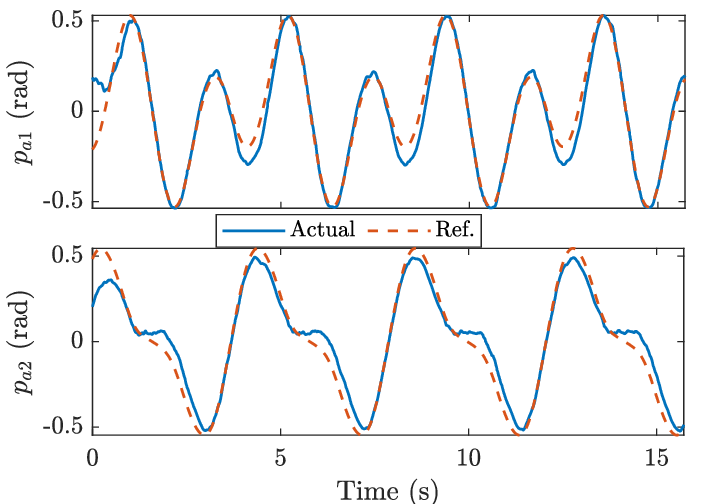}}
	\hspace{-1mm}
	\subfigure[]{
		\label{Fig_Leg_Traj_EIC_P}
		\includegraphics[width=5.35cm]{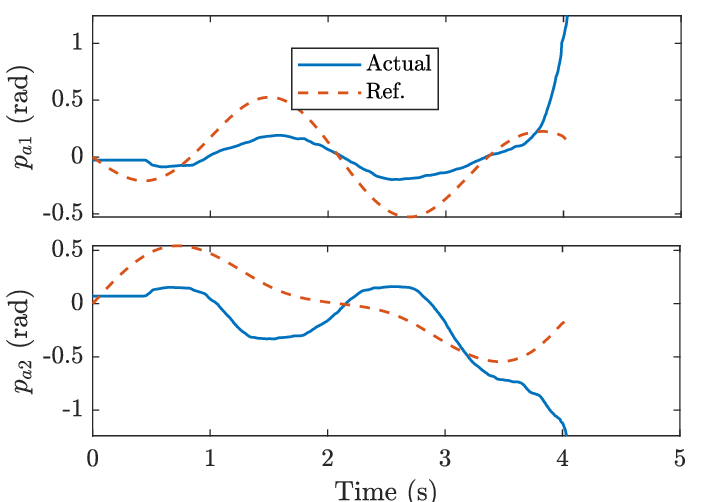}}
	\hspace{-1mm}
	\subfigure[]{
		\label{Fig_Leg_Error_Bound}
		\includegraphics[width=5.35cm]{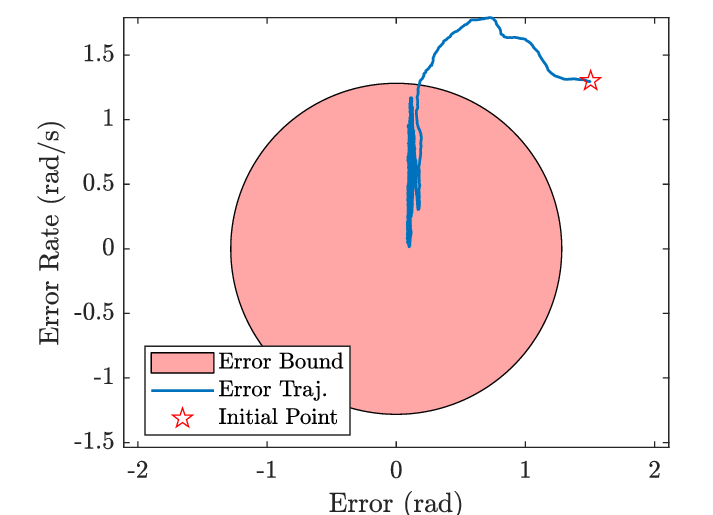}}
	\caption{Experiment results with the underactuated robotic leg. (a) Motion profiles and (b) tracking errors under the PEIC-based control. (c) Motion profiles under the EIC-based control. (d) Motion profiles in the new coordinate $\bs{p}_a$ under the PEIC-based control. (e) Motion profile $\bs{p}_a$ under the EIC-based control. (f) Error trajectory in the $\norm{\bm e_q}$-$\norm{\dot{\bm e}_q}$ plane.}
	\label{Fig_Leg}
\vspace{-4mm}
\end{figure*}

We next use a 3-DOF robotic leg ($n=2, m=1$) to demonstrate the proposed control design. The control implementation was at $200$~Hz through ROS (Robot Operating System) at a Linux real-time system machine. The nominal model is
\begin{equation*}
  \bar{\bm D} = \begin{bmatrix}
        0.15 & 0.025\c_2 & 0.025\c_3 \\
        0.025\c_2 & 0.15 & 0.05\c_{23} \\
        0.025\c_3 & 0.05\c_{23} & 0.1
      \end{bmatrix},~
  \bar{\bm H}=\begin{bmatrix}
      0 \\
      0.2\c_2\\
      0.1\s_3
    \end{bmatrix},
\end{equation*}
where $\c_{ij}=\cos(\theta_i-\theta_j)$. We apply an open-loop control (combination of sine wave torque) to excite the system and obtain the training data. The control gains were $k_{p1} = 15.0\bm I_2+20\bm \Sigma_a, k_{d1} = 3\bm I_2+10\bm \Sigma_a, k_{p2} = 25+20\Sigma_u, k_{d2} = 5.5+10\Sigma_u$. The reference trajectory was $\theta_1^d=0.5\sin t$, $\theta_2^d=0.4\sin 3t$~rad. We chose $q_{aa}=\theta_1$ and $q_{au}=\theta_2$.

Fig.~\ref{Fig_Leg} shows the experimental results. Under the proposed control, the system followed the given reference trajectory closely and the third link was balanced around BEM as shown in Fig.~\ref{Fig_Leg_Traj}. In Fig.~\ref{Fig_Leg_Error}, the tracking error of joint $\theta_1$ is between $-0.05$~rad to $0.05$~rad, while the tracking error of joint $\theta_2$ is between $-0.1$~rad to $0.1$~rad. Fig.~\ref{Fig_Leg_Traj_EIC} shows the results under the regular EIC-based control and it is clear that the system became unstable. The motion of the actuated coordinate in the new coordinate $\bs{p}_a$ is shown in Figs.~\ref{Fig_Leg_Traj_P} and~\ref{Fig_Leg_Traj_EIC_P} and $p_{a2}$ represents the uncontrolled motion variable. Though $p_{a1}$ followed the reference, the $p_{a2}$ showed a large error due to the lack of control. Fig.~\ref{Fig_Leg_Error_Bound} shows the estimated error bound and it is clear that the tracking error entered and remained inside the bounded area. The above results confirmed that under the proposed control, the uncontrolled motion is eliminated and the simultaneously tracking and balance control property of EIC-based control is preserved.

\section{Conclusion}
\label{Sec_Conclusion}

This paper proposed a learning-based controller for underactuated balance robots. The proposed control is an extension of the external and internal convertible form control (EIC-based control). The EIC-based control aims to achieve tracking and balance simultaneously. However, we showed that there exists uncontrolled motion, which can cause the system unstable. We identified the conditions under which the uncontrolled motion happened and also proposed the GP-enhanced EIC-based control. The proposed new robot control preserved the structured design of the EIC-based control and achieved tracking and balance tasks. We tested the the new control design on two experimental platforms and confirmed that stability and balance can be guaranteed.

\bibliographystyle{IEEEtran}
\bibliography{HanRef}
\end{document}